\documentclass[letterpaper, 10 pt, conference]{ieeeconf}

\IEEEoverridecommandlockouts
\usepackage[utf8]{inputenc}
\usepackage{graphicx}
\usepackage{amssymb}
\usepackage{amsmath}
\usepackage{amsfonts}
\usepackage{xcolor}
\usepackage{cite}
\let\labelindent\relax
\usepackage{enumitem}
\usepackage{caption}

\newcommand{\eg}{\textit{e.g.}}
\newcommand{\ie}{\textit{i.e.}}

\title{\LARGE \bf
Asynchronous Task Plan Refinement for Multi-Robot Task and Motion Planning}

\author{Yoonchang Sung$^{1}$, Rahul Shome$^{2}$, and Peter Stone$^{3}$% <-this % stops a space
\thanks{$^{1}$The University of Texas at Austin, Department of Computer Science
{\tt\small yooncs8@cs.utexas.edu}}%
\thanks{$^{2}$The Australian National University, School of Computing
{\tt\small rahul.shome@anu.edu.au}}%
\thanks{$^{3}$The University of Texas at Austin, Department of Computer Science and Sony AI
{\tt\small pstone@cs.utexas.edu}}%
}

\begin{document}

\maketitle
\thispagestyle{empty}
\pagestyle{empty}

%%%%%%%%%%%%%%%%%%%%%%%%%%%%%%%%%%%%%%%%%%%%%%%%%%%%%%%%%%%%%%%%%%%%%%%%%%%%%%%%
\begin{abstract}

This paper explores general multi-robot task and motion planning, where multiple robots in close proximity manipulate objects while satisfying constraints and a given goal. In particular, we formulate the plan refinement problem—which, given a task plan, finds valid assignments of variables corresponding to solution trajectories—as a hybrid constraint satisfaction problem. The proposed algorithm follows several design principles that yield the following features: (1) efficient solution finding due to sequential heuristics and implicit time and roadmap representations, and (2) maximized feasible solution space obtained by introducing minimally necessary coordination-induced constraints and not relying on prevalent simplifications that exist in the literature. The evaluation results demonstrate the planning efficiency of the proposed algorithm, outperforming the synchronous approach in terms of makespan.

\end{abstract}

%%%%%%%%%%%%%%%%%%%%%%%%%%%%%%%%%%%%%%%%%%%%%%%%%%%%%%%%%%%%%%%%%%%%%%%%%%%%%%%%

\section{Introduction}\label{sec:intro}

Developing multi-robot systems to achieve a desired goal while interacting with objects in the world requires integrated reasoning about task sequencing, task allocation, and motion planning. Task and motion planning (TAMP~\cite{garrett2021integrated}) jointly addresses the search for a sequence of discrete symbolic actions, the selection of which object to manipulate, and the assignment of continuous values to actions, determining how to execute those actions. However, the TAMP literature has predominantly focused on single-robot problems. 

Another closely related topic is multi-robot motion planning~\cite{stern2019multi,madridano2021trajectory}, which aims to find collision-free paths for multiple robots. In this context, objects are not considered for manipulation but rather are treated as obstacles. Additionally, multi-robot motion planning typically addresses individual motion planning problems, unlike TAMP where a sequence of motion planning problems is considered. The objective of this work is to develop a general-purpose multi-robot TAMP (MR-TAMP) framework that inherits challenges from both of these perspectives. 

In existing MR-TAMP research, two prevalent simplifications are the \emph{pre-discretization} of the search space~\cite{brown2020optimal,chen2021decentralized} and \emph{synchronous actions}~\cite{shome2021synchronized,pan2021general,zhang2022mip,ahn2022coordination}, where robots simultaneously initiate and complete action execution. While these assumptions simplify algorithm design, they can significantly diminish the space of feasible solutions, potentially preventing the solution of certain feasible problems and reducing the diversity of available solution paths.
% Introducing these assumptions presents a critical limitation, as formerly feasible problems may no longer remain feasible, and the diversity of solution paths can be lost.

In this work, our goal is to formulate MR-TAMP problems that maximize the feasible solution space by avoiding both of these simplifications. This approach can be viewed as an extension of the TAMP formulation to MR-TAMP, introducing only the necessary constraints arising from multi-robot coordination. The formulation essentially represents a \emph{hybrid} constraint satisfaction problem (H-CSP~\cite{lozano2014constraint,garrett2021integrated}), incorporating both discrete and continuous variables.

To achieve this goal, we address a specific aspect in this work, referred to as the \emph{refinement} problem. When a task plan is provided, specifying the sequences of object manipulations for all robots, the objective of the refinement problem is to assign values to all continuous variables that meet the constraints, in order to find solution paths that the robots can execute. This direction shows promise, as we can seamlessly harness state-of-the-art multi-agent task planners from the AI planning community~\cite{de2009introduction,torreno2017cooperative} when developing the complete framework in the future. 
% Given that a task plan is assumed to be provided in this work, the proposed refinement problem is closely related to multi-modal motion planning~\cite{hauser2010multi} that does not explicitly involve task planning.

\begin{figure}[ht]
\centering
\includegraphics[width=0.45\textwidth]{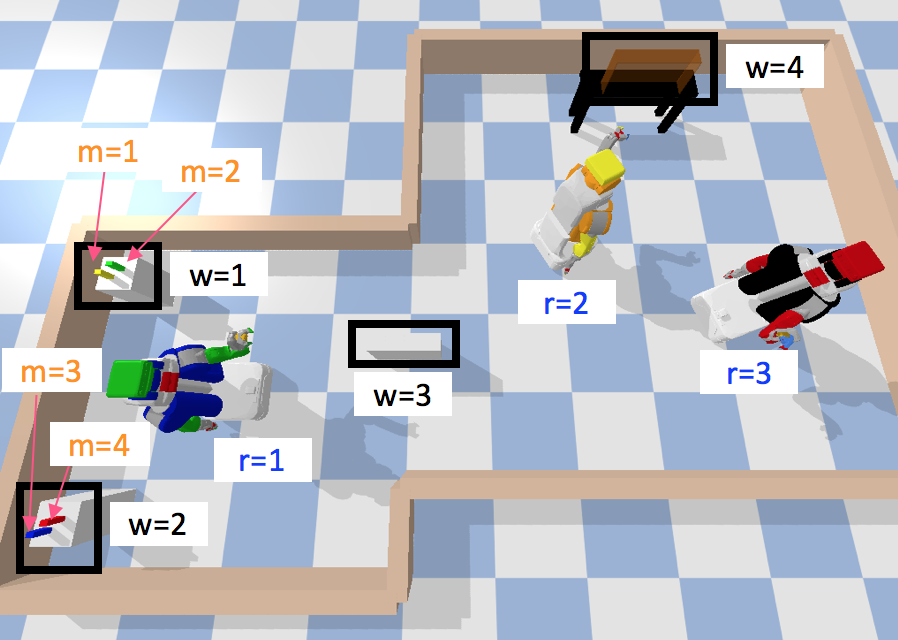}
\caption{\small Example MR-TAMP task showing the initial state. Robots ($\{r\}_{r=1}^3$), movable objects ($\{m\}_{m=1}^4$), and workspace regions ($\{w\}_{w=1}^4$) are depicted in the figure, while fixed objects, corresponding to walls, shelves, table, and cabinet, are not shown. The goal is to move all movable objects from their initial locations to the cabinet (\ie, workspace region $4$). The given task involves three robots such that robot $1$ moves all movable objects to workspace region $3$, and robots $2$ and $3$ move them to workspace region $4$.
}
\label{fig:task}
\end{figure}

Figure~\ref{fig:task} illustrates the type of task we address, wherein multiple mobile manipulators operate in close proximity, involving multi-step manipulations such as picking up and placing multiple objects. The process of solving the proposed refinement problem, which aims to satisfy the given task plan and constraints, reveals specific grasp poses, placements, motions, and action scheduling for the robots.

Our main contributions can be summarized as follows: (1) the introduction of a general problem formulation for MR-TAMP that is inherently asynchronous and does not require complex scheduling, (2) the identification of fundamental challenges raised by this problem, and (3) the proposal of a search algorithm that incorporates promising heuristics. 
% \todo{Summary of experimental results}.
% \rs{The scheduling/asynchronous point might be worth mentioning early on.}{}

\section{The MR-TAMP Refinement Problem}\label{sec:prob}

\subsection{Notations and assumptions}
\label{subsec:notation}
Consider $R$ robots, indexed as $\{r\}_{r=1}^{R}$, manipulating objects to achieve a goal in a 3D workspace. The workspace consists of $M$ movable objects, such as cups and plates, indexed as $\{m\}_{m=1}^M$ and $F$ fixed objects, such as tables and shelves, indexed as $\{f\}_{f=1}^F$. We denote $W$ workspace regions as $\{w\}_{w=1}^W$, where movable objects can be placed, such as the surface of the table and the space on the shelf, inspired by the work~\cite{berenson2011task}.

While our framework is not necessarily restricted to homogeneous robots (\ie, robots with the same shapes, degrees of freedom, and abilities to move and manipulate), in this paper, we consider homogeneous robots for the sake of notational convenience. Each robot $r$ operates in a $d$-dimensional configuration space whose configuration is represented as $q_r\in\mathcal{C}_r\subset\mathbb{R}^d$. The pose of a movable object $m$ is denoted as $p_m\in\mathcal{P}_m\subset \mathit{SE}(3)$. Then, the composite configuration space for all robots and movable objects becomes $\mathcal{C}=\prod_{r=1}^R\mathcal{C}_r\times\prod_{m=1}^M\mathcal{P}_m$. We denote the free space of the composite configuration space as $\mathcal{C}^\textsc{f}$, which represents all possible configurations of robots and movable objects that are positioned stably and do not collide with each other and with fixed objects. Correspondingly, the obstacle space is defined as $\mathcal{C}^\textsc{o}=\mathcal{C}\setminus\mathcal{C}^\textsc{f}$.

We assume quasi-static dynamics in the world, which implies that movable objects remain stable after being manipulated by robots. Additionally, we assume that each movable object can be manipulated by a single robot. Furthermore, we assume deterministic transition effects, full observability, and lossless communication among robots. While our focus in this work is on pick-and-place tasks where geometric constraints are of major concern, our ultimate aim is to position this work as a foundational framework in MR-TAMP that can effectively address a wider range of practical challenges in the future, including those that relax the assumptions mentioned in this paragraph.

\subsection{Mode-based abstract actions}
\label{subsec:abs_action}
We employ the notion of a \emph{mode}~\cite{simeon2002path,hauser2010multi,garrett2018ffrob,vega2020asymptotically}, denoted by $\sigma$, which specifies a constraint submanifold of $\mathcal{C}^\textsc{f}$, to define actions. These modes are determined by the contact points between the robot and the movable object (\eg, robot $r$ grasping movable object $m$), while the remaining objects remain stationary. We consider two types of modes: a \emph{transit mode} $\sigma^\textsc{s}$ where a robot moves with an empty hand, and a \emph{transfer mode} $\sigma^\textsc{f}$ where a robot moves while holding a movable object. The transition between two adjacent modes can be facilitated through a \emph{transition configuration}, which represents the robot's grasping or placing configuration.

We define the \emph{abstract action} based on these two modes. Let the abstract action be $a=\big\{\sigma^{\textsc{s}}(r, m, w, w^\prime),$ $\sigma^{\textsc{f}}(r, m, w, w^\prime)\big\}$. $\sigma^{\textsc{s}}(r, m, w, w^\prime)$ indicates that robot $r$ moves from workspace region $w$ to another workspace region $w^\prime$ with an empty hand in order to grasp movable object $m$ located in $w^\prime$. $\sigma^{\textsc{f}}(r, m, w, w^\prime)$ indicates that robot $r$, while already grasping movable object $m$ in workspace region $w$, moves and places it in another workspace region $w^\prime$. These actions are still abstract because continuous parameters, such as robot configurations $\{q_r\}_{r=1}^R$ and object poses $\{p_m\}_{m=1}^M$, are not yet specified. Abstract actions may encompass both arm and base motions, as illustrated in Figure~\ref{fig:task}.

% \rs{Something you should distinguish here is whether your framework can handle any abstract actions beyond manipulation? This might affect how strongly you want to claim TAMP versus Multi-Robot Manipulation in the intro/contribution.}{}

% Unparameterized plan skeletons~\cite{lozano2014constraint}

\subsection{H-CSP for refinement}
\label{subsec:hcsp}
We formulate the refinement of abstract actions into fully specified actions that robots can execute as an H-CSP problem. This problem involves assigning values from the domains of variables while ensuring that the assigned values do not violate any constraints. The variable set is defined as $\mathcal{V}=\big\{\{v_r^q\}_{r=1}^R, \{v_r^g\}_{r=1}^R, \{v_m^p\}_{m=1}^M\big\}$, where $v_r^q$ is a transition configuration variable for robot $r$, $v_r^g$ is a grasp variable for robot $r$, and $v_m^p$ is a pose variable for movable object $m$. The domains for these variables are defined as follows: for $v_r^q$, $\mathcal{D}_r^q=\mathcal{C}_r$; for $v_r^g$, $\mathcal{D}_r^g=\cup_{m=1}^M\mathcal{G}_{r,m}$; and for $v_m^p$, $\mathcal{D}_m^p=\mathcal{P}_m$. Here, $\mathcal{G}_{r,m}\ni g_{r,m}=(r, m, \gamma_{r,m})$ indicates that robot $r$ grasps movable object $m$ with a relative transformation $\gamma_{r,m}$ between the pose of the robot $r$'s end-effector and the pose of the object $p_m$.
% Note that an assigned value $q_i$ for its variable $v_i$ also specifies the grasp pose of robot $r_i$ for the corresponding movable object, as $v_i$ corresponds to a transition configuration. 
The abstract actions are associated with variables as goal variables, where $\sigma^\textsc{s}(r, m, w, w^\prime)$ includes $v_r^q$ and $v_r^g$, while $\sigma^\textsc{f}(r, m, w, w^\prime)$ includes $v_r^q$ and $v_m^p$.

We present the mode-specific constraints, which are parameterized, that the assigned values must satisfy as follows:
\begin{itemize}[leftmargin=*]
\item $\texttt{Motion}\big(q_r, q^\prime_r; G_r=(V_r,E_r)\big)$ represents a reachability constraint for robot $r$ from a start configuration $q_r$ to a goal configuration $q^\prime_r$. This constraint is verified by applying existing motion planners. In our case, we utilize probabilistic roadmap (PRM~\cite{kavraki1996probabilistic}) planners, which are a well-known sampling-based motion planner. These planners generate a roadmap data structure $G_r=(V_r,E_r)$, where the vertex set $V_r$ consists of robot $r$'s configurations, and the edge set $E_r$ consists of edges (often straight lines in $C_r$) that connect pairs of vertices in $V_r$. The PRM planners assist in finding a path that connects the start configuration $q_r$ and the goal configuration $q^\prime_r$.

\item $\texttt{CFree}\big(\{q_r\}_{r=1}^R, \{p_m\}_{m=1}^M, \{f\}_{f=1}^F \big)$ ensures that there are no pairwise collisions among robots at $\{q_r\}_{r=1}^R$ configurations, movable objects at $\{p_m\}_{m=1}^M$ poses, and fixed objects $\{f\}_{f=1}^F$. Eventually, the tensor product of individual roadmaps $\{G_r\}_{r=1}^R$, as shown in the $\texttt{Motion}$ constraint, must find $R$ paths corresponding to $R$ robots that satisfy this $\texttt{CFree}$ constraint.
% The tensor-product roadmap $G$, which will be formally defined in Section~?, is the tensor product of individual roadmaps $\{G_r\}_{r=1}^R$, shown in the $\texttt{Motion}$ constraint. Every vertex and edge in $G$ 

\item $\texttt{Kin}(q_r, g_{r,m}, p_m)$ guarantees the existence of a kinematic solution for grasping movable object $m$ at pose $p_m$ with grasp $g_{r,m}$ and robot $r$'s configuration $q_r$.

\item $\texttt{Grasp}(g_{r,m},p_m)$ represents a graspability constraint, indicating that a movable object at pose $p_m$ can be grasped with grasp $g_{r,m}$.

\item $\texttt{Hold}(r,m)$ ensures that movable object $m$ is securely attached to the hand of robot $r$. When this constraint is activated, it affects other constraints in the following manner. In the $\texttt{CFree}$ constraint, the pose $p_m$ of movable object $m$ is no longer considered directly, but can be computed based on grasp $g_{r,m}$ and robot $r$'s configuration $q_r$. Additionally, the collision detection between robot $r$ and movable object $m$ is no longer considered in the $\texttt{CFree}$ constraint. Furthermore, it prevents the activation of $\texttt{Grasp}$ constraints for other robots besides $r$, ensuring that the same movable object cannot be grasped by multiple robots while it is already being held.

\item $\texttt{Contain}(m,w)$ constrains that movable object $m$ is stably placed within workspace region $w$.
\end{itemize}

Among the constraints, $\texttt{Motion}$, $\texttt{CFree}$, and $\texttt{Kin}$ are always applied to both types of abstract actions, $\sigma^\textsc{s}$ and $\sigma^\textsc{f}$. $\texttt{Grasp}$ and $\texttt{Contain}$ constraints serve as goals within the abstract actions. The constraints applied for each type of abstract action are presented as follows:
\begin{itemize}
\item $\sigma^\textsc{s}$: $\texttt{Motion}$, $\texttt{CFree}$, $\texttt{Kin}$, and $\texttt{Grasp}$.
\item $\sigma^\textsc{f}$: $\texttt{Motion}$, $\texttt{CFree}$, $\texttt{Kin}$, $\texttt{Hold}$, and $\texttt{Contain}$.
\end{itemize}

Note that we do not introduce a constraint enforcing the synchronous start and end of abstract actions for all robots, characterizing the synchronous approach. Therefore, our formulation strictly generalizes the synchronous formulation.

\subsection{The proposed problem}
\label{subsec:prob}

In this work, we address a partial problem where \emph{ground} abstract actions for all robots are provided, which means that the arguments $r$, $m$, $w$, and $w^\prime$ are grounded in all instances of $\sigma^\textsc{s}$ and $\sigma^\textsc{f}$, as well as the ordering among abstract actions. However, we still need to assign values to the variables of the corresponding abstract actions that satisfy the constraints specified in Section~\ref{subsec:hcsp}. This particular approach is referred to as the \emph{sequence-before-satisfy} strategy in the TAMP literature~\cite{garrett2021integrated}, and our focus is on addressing the satisfy part, or refinement, assuming that sequencing is given.

Specifically, we are provided with a tuple $\big<\{a_r^{A_r}\}_{r=1}^{R}, \prec\big>$, where $a_r^{A_r}$ represents a set of abstract actions for robot $r$, and $A_r$ is an index set specific to robot $r$, allowing robots to have different cardinalities of abstract actions. $\prec$ is a set of ordering constraints that determine the sequencing of the provided abstract actions.

It is important to note that these ordering constraints can apply not only to abstract actions of the same robot but also to abstract actions of different robots. For instance, if movable object $m$ is initially placed in workspace region $w$, then the refinement of $\sigma^\textsc{s}(r,m,w^\prime,w^{\prime\prime})$ for robot $r$ cannot be carried out until another robot $r^\prime$ executes $\sigma^\textsc{s}(r^\prime,m,w,w^\prime)$, as the movable object $m$ is not yet located within the workspace region $w^\prime$. 
% \rs{You might want to refer to task space regions as an inspiration for formalizing w}{Task Space Regions: A framework for pose-constrained manipulation planning by Berenson et al}

Furthermore, $\prec$ does not specify the ordering between every pair of abstract actions from $\{a_r^{A_r}\}_{r=1}^{R}$. $\prec$ is \emph{minimally} given in the sense that it only specifies the sequence of workspace regions where each movable object is placed. Any orderings that require geometric reasoning are not included and must be determined by solving the refinement problem. For instance, suppose workspace region $w$ has limited space. In that case, robot $r$ can only feasibly place movable object $m$ in workspace region $w$ (\eg, $\sigma^\textsc{f}(r,m,w^\prime,w)$) after another robot $r^\prime$ removes another movable object $m^\prime$ from the same workspace region (\eg, $\sigma^\textsc{f}(r^\prime,m^\prime,w,w^\prime)$), creating empty space in workspace region $w$.

Let $s_0=\big((q_r)_{r=1}^R, (p_m)_{m=1}^M\big)$ represent the initial state, specifying the initial configurations of all robots and the initial poses of all movable objects. The refinement problem is then defined as follows: given a tuple $\big<\{a_r^{A_r}\}_{r=1}^{R}, \prec, s_0\big>$, the goal is to find valid assignments of variables defined in Section~\ref{subsec:hcsp} for all abstract actions $\{a_r^{A_r}\}_{r=1}^{R}$, potentially introducing additional ordering constraints while respecting the given ordering constraints $\prec$ and the mode-specific constraints. 

\section{Algorithm}\label{sec:alg}

Solving the proposed problem while respecting all the constraints simultaneously is highly challenging, as even a single robot TAMP problem is known to be intractable (\ie, PSPACE-hard~\cite{canny1988complexity}). Additionally, explicitly constructing a composite roadmap from $\{G_r\}_{r=1}^R$ is computationally expensive, especially considering the exponential increase in the number of samples required by the motion planner (such as PRM in our case) to cover the composite configuration space for all robots (\ie, $\prod_{r=1}^R C_r$). Moreover, the path for an abstract action and its length can only be determined after it has been computed by the motion planner, making it difficult to anticipate in advance when a robot will place a movable object. Consequently, it is challenging to identify when the $\texttt{CFree}$ constraints are affected by the $\texttt{Hold}$ constraints without evaluating all the relevant $\texttt{Motion}$ constraints.

\subsection{Overall framework}
\label{subsec:framework}
We propose a heuristic-based search algorithm to efficiently solve the refinement problem, incorporating the following four principles. 

(1) \textbf{Least commitment}: We follow the \emph{least commitment} principle~\cite{weld1994introduction}, avoiding the introduction of additional ordering constraints unless absolutely necessary. This approach increases the size of the feasible solution space, leading to a more diverse set of solutions. 

\begin{figure}[ht]
\centering
\includegraphics[width=0.45\textwidth]{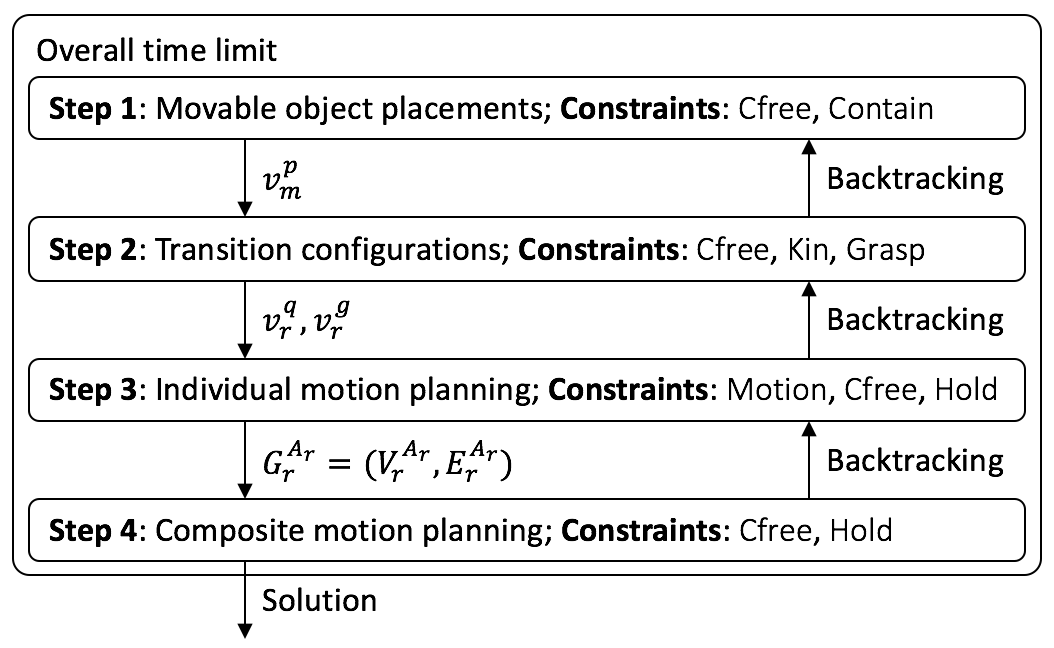}
\caption{The overall framework.}
\label{fig:framework}
\end{figure}

(2) \textbf{Sequential heuristics}: Instead of solving the problem in one step, we decompose it into a sequence of subproblems. We relax the problem by neglecting some of the mode-specific constraints, creating a relaxed problem that serves as a necessary condition for the subsequent problem in the sequence. The first subproblem is the most relaxed, and as we progress through the sequence, the neglected constraints are reintroduced incrementally.  Additionally, the relaxed problem provides heuristics for guiding the search in the next subproblem. This decomposition approach is appealing because it can efficiently find a solution if one exists or effectively detect infeasibility in the early stage of the sequence. The flow chart illustrating this process is depicted in Figure~\ref{fig:framework}. 

(3) \textbf{Implicit time representation}: Unlike many existing multi-robot task planning or TAMP approaches that explicitly represent time for temporal planning, our formulation and algorithm do not require explicit time representation. This approach avoids the complexity of introducing a scheduling problem and aligns with the observation made by Boutilier and Brafman~\cite{boutilier2001partial} that explicit time representation is not always necessary. In our approach, time is implicitly revealed as a byproduct of solving the refinement problem.

(4) \textbf{Implicit composite roadmap construction}: As mentioned at the beginning of this section, explicit construction of a composite roadmap from $\{G_r\}_{r=1}^R$ is impractical. Instead, we employ the concept of implicit composite roadmap construction, referred to as \emph{subdimensional expansion} in the literature~\cite{wagner2015subdimensional,solovey2016finding,shome2020drrt}. This approach involves generating individual roadmaps for each robot independently, ignoring collisions with other robots. These individual roadmaps are then combined in a manner that takes into account robot-robot collisions. The resulting composite roadmap consists only of explored vertices and edges.

In the following subsections, we present each component of the algorithm depicted in Figure~\ref{fig:framework}.

\subsection{Movable object placements}
\label{subsec:place}

In this step, we relax most of the mode-specific constraints and retain only the $\texttt{CFree}$ and $\texttt{Contain}$ constraints in all abstract actions. Moreover, in the $\texttt{CFree}$ constraint, we disregard the robot configurations from the argument, resulting in $\texttt{CFree}\big(\{p_m\}_{m=1}^M, \{f\}_{f=1}^F\big)$. This step can be seen as the teleportation of movable objects from one workspace region to another, excluding any robot involvement. The objective is to find valid assignments for all the relevant pose variables $v_m^p$ present in the given abstract actions $\{a_r^{A_r}\}_{r=1}^{R}$ to satisfy the $\texttt{Contain}$ constraint and, if necessary, introduce additional ordering constraints to resolve the $\texttt{CFree}$ constraint with other movable objects.

This subproblem can be effectively solved by further decomposing it into multiple workspace region-specific problems since placements in one workspace region are completely independent of those in other regions. Let's consider a specific workspace region $w$ present in the given abstract actions $\{a_r^{A_r}\}_{r=1}^{R}$; we apply the same procedure to other relevant workspaces. For workspace region $w$, we find a set of sequences that specify the ordering of addition (or placement) and removal operations for each relevant movable object. Note that this sequence set can be derived entirely from $\big<\{a_r^{A_r}\}_{r=1}^{R}, \prec\big>$, and that each sequence consists of an alternating sequence of addition and removal operations.

From the sequence set in workspace region $w$, we determine pose variable assignments for the subset of sequences that involve addition operations. We employ a sampling strategy by uniformly drawing a predetermined number of placement samples in workspace region $w$ for each movable object in the subset. 

To avoid unnecessary introduction of additional ordering constraints, we make the following observation: If we can solve the most constrained problem, where the $\texttt{CFree}$ constraint is applied to all movable objects already located in workspace region $w$ and those that will be added, no additional ordering constraints will be necessary. Only when the $\texttt{CFree}$ constraint is violated for some movable objects, do we introduce additional ordering constraints, ensuring that one movable object is added after another is removed. This observation is based on the idea that in spacious workspace regions, the $\texttt{CFree}$ constraint is mostly satisfied without the need for additional ordering constraints. However, in tiny workspace regions, many ordering constraints may be required.

In this step, for each abstract action under consideration, we store information about the movable objects from $\{m\}_{m=1}^M$ and fixed objects from $\{f\}_{f=1}^F$ involved in evaluating collisions with the corresponding movable object. This cached information will be utilized in the subsequent steps.

If no valid assignments can be found even after evaluating all possible combinations of the predetermined number of placement samples, we have two options. First, we can stop the process and declare the problem as infeasible. In this case, the next steps do not need to be attempted, as they rely on finding valid assignments in this subproblem. Alternatively, we can choose to draw more samples until a predetermined time limit is reached.

\subsection{Transition configurations}
\label{subsec:transit}

After obtaining valid assignments for all relevant pose variables associated with abstract actions $\{a_r^{A_r}\}_{r=1}^{R}$, our next step is to find valid assignments for all relevant transition configuration variables $v_r^q$ and grasp variables $v_r^g$. However, in this process, we continue to disregard certain mode-specific constraints, such as $\texttt{Motion}$ and $\texttt{Hold}$, as well as the presence of other robots. Instead, we focus on considering the $\texttt{CFree}$, $\texttt{Kin}$, and $\texttt{Grasp}$ constraints. This step aims to identify feasible transition configurations and grasps for all given abstract actions $\{a_r^{A_r}\}_{r=1}^{R}$ that are compatible with the movable object poses obtained in the previous step. 
% By combining the object poses with the determined goal configurations and grasps, we can fully specify all the transition configurations present in the given abstract actions $\{a_r^{A_r}\}_{r=1}^{R}$.

We no longer need to take the workspace region-specific approach as in the previous step. Instead, we address this subproblem for each pair of abstract actions of the same robot consisting of $\sigma^\textsc{s}$ and $\sigma^\textsc{f}$ sequenced by the ordering constraints $\prec$. Let's consider the sequential abstract actions corresponding to robot $r$, denoted as $\sigma^\textsc{s}(r,m,w,w^\prime)$ and $\sigma^\textsc{f}(r,m,w^\prime,w^{\prime\prime})$. These abstract actions indicate that robot $r$ grasps movable object $m$ in workspace region $w^\prime$ and moves to workspace region $w^{\prime\prime}$ to place the object there. The same rule is applied to all other pairs of abstract actions of the same robot sequenced by the ordering constraints $\prec$.

Instead of considering $\{q_r\}_{r=1}^R$ as arguments in the $\texttt{CFree}$ constraint, we only consider robot $r$'s configuration $q_r$, ignoring other robots. As for the remaining object-related arguments, we retrieve the collision information cached in the previous step, which indicates which objects must be considered for collision checking. Since collisions among objects have already been confirmed in the previous step, we only assess collisions between robot $r$ and the relevant objects using the $\texttt{CFree}$ constraint.

Since the $\texttt{Grasp}$ constraint is associated with the mode $\sigma^\textsc{s}$, we first find a valid assignment for the grasp variable $v_r^g$ corresponding to the abstract action $\sigma^\textsc{s}$ by sampling a predetermined number of grasps. Once a valid grasp is found, we compute $q_r$ with respect to the grasp $g_{r,m}$ using the $\texttt{Kin}$ constraint. In the case of a mobile manipulator, as used in our experiments, computing $q_r$ involves determining a base pose and subsequently solving an inverse kinematic problem (\ie, $\texttt{Kin}$) to verify reachability to grasp $g_{r,m}$~\cite{diankov2010automated}. This computed $q_r$ is for the abstract action $\sigma^\textsc{s}$. Similarly, the same grasp $g_{r,m}$ is used to find another $q^\prime_r$ for the corresponding abstract action $\sigma^\textsc{f}$. The computed configurations $q_r$ and $q^\prime_r$ are then used in their respective $\texttt{CFree}$ constraints to ensure collision-free transition configurations. 

If valid transition configurations can be found for all the abstract actions $\{a_r^{A_r}\}_{r=1}^{R}$ from the set of possible grasp samples, we can proceed to the next step. However, if valid transition configurations cannot be found, we have three options. First, we can choose to stop the process as explained in the previous step, indicating that a solution cannot be found. Second, we can backtrack to the previous step and explore unevaluated combinations of placement samples to potentially find valid transition configurations. To improve efficiency, we can also inform the previous step about the cause of failure, allowing suitable ordering constraints to be added and prevent the same failures in future attempts. Lastly, we can increase the number of grasp samples and reevaluate this step to improve the chances of finding valid transition configurations.

\subsection{Individual motion planning}
\label{subsec:individual}

Even after obtaining feasible transition configurations, as mentioned in the fourth principle, solving for paths of all robots simultaneously by explicitly constructing a composite roadmap is a challenging task. To address this complexity, we leverage the discrete RRT (dRRT~\cite{solovey2016finding,shome2020drrt}) algorithm, which is built upon the subdimensional expansion concept. The dRRT algorithm is specifically designed for solving \emph{single-modal} motion planning problems involving multiple robots. In our algorithm, we extend the capabilities of dRRT in two aspects: (1) individual motion planning is generalized to multi-modal motion planning, considering multiple abstract actions, and (2) our algorithm accommodates robots holding objects, which affects the collision-checking process.

In this step, we focus on considering the $\texttt{Motion}$ and $\texttt{Hold}$ constraints, given feasible transition configurations. During individual motion planning, we still disregard the presence of other robots. Furthermore, we assume that all movable objects, except for the one held by the corresponding robot, have been placed in their respective workspace regions, as determined in the movable object placement step. As a result, the $\texttt{CFree}$ constraint still includes the same arguments as in the previous transition configuration step. However, the $\texttt{Hold}$ constraint allows for collision between the robot and the movable object it holds.

Unlike the previous steps, we decompose this subproblem into multiple individual motion planning problems. Specifically, we can find a sequence of abstract actions for each robot from $\big<\{a_r^{A_r}\}_{r=1}^{R}, \prec\big>$ and apply PRM to each abstract action in the sequence. In this case, the transition configurations serve as start and goal configurations, and we generate a predetermined number of samples in the respective configuration space $C_r$. Throughout this process, we apply the $\texttt{CFree}$ and $\texttt{Hold}$ constraints as mentioned before. This subproblem can be seen as the verification of reachability from the start transition configuration of the first abstract action to the goal transition configuration of the last abstract action. 
% To effectively address this, we apply PRM from the first abstract action to the last abstract action.

If valid individual paths can be found for all robots, we can proceed to the last step. Otherwise, we have the same three options as in the transition configuration step.

\subsection{Composite motion planning}
\label{subsec:composite}

We are now ready to consider all the intact mode-specific constraints introduced in Section~\ref{subsec:hcsp} by merging the individual paths obtained from the previous step. This step involves constructing a tensor-product roadmap from individual roadmaps $\{G_r^{A_r}=(V_r^{A_r},E_r^{A_r})\}_{r=1}^R$, where $A_r$ is the abstract action index set for robot $r$. We denote the resulting tensor-product roadmap as $G=(V,E)$. In $G$, the set of vertices $V$ is the Cartesian product of the vertices from $\{G_r^{A_r}\}_{r=1}^R$, represented as $V=\{(v_1, ..., v_r, ..., v_R)|\forall r\ v_r\in V_r^{A_r}\}$. The set of edges $E$ is defined as $E=\big\{\big((v_1, ..., v_r, ..., v_R), (v_1^\prime, ..., v_r^\prime, ..., v_R^\prime)\big)\big|\forall i\ \exists (v_r,v_r^\prime)\  \big((v_r,v_r^\prime)\in E_r^{A_r} \vee v_r=v_r^\prime \big)\big\}$. Note that in $E$, the condition $v_r=v_r^\prime$ allows some robots to remain stationary. However, since robot-robot collisions and collisions between robots and movable objects held by other robots were not considered in the $\texttt{CFree}$ constraint in the previous steps, some edges in $E$ may contain collision paths among robots.

Due to limited space, we provide a brief explanation of how dRRT works and how we modify it for our problem. For detailed explanations, please refer to the works~\cite{solovey2016finding,shome2020drrt}. dRRT is based on RRT~\cite{lavalle1998rapidly} and serves as the underlying framework for constructing the composite search graph $G$. dRRT incrementally builds $G$ by sampling configurations in the composite configuration space $\prod_{r=1}^R C_r$ and connecting them using an oracle function that searches for neighboring vertices. The oracle function finds the nearest neighbor vertex $v_r$ and another neighbor vertex $v_r^\prime$ within the individual roadmap $G_r^{A_r}$ for a given sampled configuration. During the composite search, the intact $\texttt{CFree}$ and $\texttt{Hold}$ constraints, as explained in Section~\ref{subsec:hcsp}, are used to ensure collision-free and object-holding paths.

During the composite search, when the goal configuration (\ie, transition configuration) of one robot's roadmap is reached, the next roadmap for the same robot is considered. The ordering constraints $\prec$ are taken into account in the composite search, ensuring that no adjacent edges connected to a goal configuration of the corresponding roadmap are used until another robot's roadmap, as determined by $\prec$, is reached.

If the modified dRRT algorithm finds a valid composite path for all robots, we declare that a solution path satisfying the mode-specific constraints and ordering constraints has been found, given the input $\big<\{a_r^{A_r}\}_{r=1}^{R}, \prec, s_0\big>$. dRRT has its own time limit, and if this limit is exceeded, we backtrack to the previous step. Additionally, we set an overall time limit for the entire process, and if this limit is exceeded, the algorithm terminates with no solution.

\section{Experiments}
\label{sec:experiment}

We perform two sets of experiments in PyBullet~\cite{coumans2021} to evaluate the performance of the proposed algorithm. (1) Ablation study: We analyze the effectiveness of decomposition by comparing planning time with merged hierarchies. (2) Comparison with the synchronous approach: We evaluate the makespan (\ie, the execution time of the last robot) of our algorithm against the synchronous method to highlight our method's ability to discover more effective solutions.

All the experiments are conducted using the task shown in Figure~\ref{fig:task}. We consider mobile manipulators as our robots with three and seven-dimensional configuration spaces for base motion and arm motion, respectively. Each abstract action consists of a sequence of three motion planning problems: base motion reaching a desired base position, arm motion grasping a target object, and arm motion returning to a home position. Base poses and grasp poses are all sampled, as is typically done in the literature~\cite{diankov2010automated,garrett2018ffrob}. Due to the limited space, we provide the details of the task, such as specifications of input tuple $\big<\{a_r^{A_r}\}_{r=1}^{R}, \prec\big>$, in the video. As the task contains $15$ abstract actions, there are a total of $45$ individual motion planning problems to solve the task. We report the results in Table~\ref{table:result}, where statistics are collected by solving the problem with $25$ different random seeds.

\begin{table}[h!]
\begin{center}
\begin{tabular}{|c||cccccc|}
\hline
\hline
Algorithms & \multicolumn{2}{c|}{Our algorithm} & \multicolumn{2}{c|}{MERGE $1\&2$} & \multicolumn{2}{c|}{MERGE $1$--$3$}\\
\hline
Planning time (s) & \multicolumn{2}{c|}{$324.7\pm40.2$} & \multicolumn{2}{c|}{$371.2\pm54.6$} & \multicolumn{2}{c|}{$-$}\\
\hline
\hline
\end{tabular}
%%%%%%
\begin{tabular}{|c||cccccc|}
Algorithms & \multicolumn{3}{c|}{Our algorithm} & \multicolumn{3}{c|}{Synchronous}\\
\hline
Makespan (simulation steps) & \multicolumn{3}{c|}{$5118.3\pm148.4$} & \multicolumn{3}{c|}{$7432.1\pm211.8$}\\
\hline
\hline
\end{tabular}
\caption{\small Experimental results. The numbers represent mean and 95$\%$ confidence interval. $-$ implies that all instances take longer than $10$ minutes to solve.}
\label{table:result}
\end{center}
\end{table}

\textbf{Ablation study}: Since the importance of the decomposition between Steps 3 and 4 is emphasized in dRRT~\cite{solovey2016finding,shome2020drrt}, we focus on the importance of decomposition among Steps 1, 2, and 3. The first ablation is to merge Steps 1 and 2 (\ie, MERGE $1\&2$), and the second one is to merge Steps 1, 2, and 3 (\ie, MERGE $1$--$3$). 

The results in the first row of Table~\ref{table:result}  indicate that useful heuristics can be found by decomposition, and thus, a solution is found quickly. MERGE $1$--$3$ takes longer than $10$ minutes in all instances due to the generation of unnecessary motion planning problems in Step 3 that do not lead to a solution. We observe some differences between our algorithm and MERGE $1\&2$, but they are not significant. This implies that, although MERGE $1\&2$ had to solve many unnecessary inverse kinematic problems, the heuristic found by Step 2 is powerful in solving the rest of the problem, as Steps 3 and 4 consume the majority of planning time.

\textbf{Comparison with the synchronous approach}:
In the synchronous approach, all robots either leave and arrive at their corresponding transition configurations at the same time or remain idle during that time period. In tasks where robots manipulate objects in the same workspace regions (\eg, all robots converge at workspace region $3$), if the planner does not find feasible transition configurations for all robots, some robots need to remain idle. Moreover, robots $2$ and $3$ can only start moving to workspace region $3$ after robot $1$ places an object there. 

Makespan results in the second row of Table~\ref{table:result} support that our asynchronous algorithm is more execution time efficient than the synchronous one, which aligns with the above observations. In any case, the synchronous approach is impractical; if one of the abstract actions requires a robot to move a long distance, all the remaining robots must wait.

\section{Related Work}
\label{sec:related}

In this section, we briefly review existing MR-TAMP research, in addition to those referred to in the introduction, which rely on pre-discretization or the synchronous approach. Various task types have been investigated, such as assembly~\cite{knepper2013ikeabot,hartmann2022long,chen2022cooperative} and clutter removal~\cite{tang2020computing}. Challenges that have not been addressed in this work are discussed in the context of MR-TAMP, including decentralized communication~\cite{chen2021decentralized} and spatial and temporal uncertainty~\cite{faroni2023optimal}.

One distinguishing feature of this work is its implicit time representation, whereas the majority of existing works~\cite{karami2021task,messing2022grstaps,chen2022cooperative,hartmann2022long} reason about time explicitly, which incurs the relatively complex overhead of task scheduling.

To solve MR-TAMP problems efficiently, various approximations have been introduced, including state space decomposition~\cite{motes2020multi,karami2021task,motes2023hypergraph} and shared space graph~\cite{umay2019integrated}. Although incorporating approximations may lead to a loss of feasibility guarantees, it is an interesting avenue for future research.

Optimization-based approaches~\cite{toussaint2017multi,hartmann2022long} have made progress in MR-TAMP by leveraging logic-geometric programming~\cite{toussaint2015logic}. The most recent work~\cite{hartmann2022long} in this direction focuses on the assembly task but still relies on explicit time representations.

% Composite and decoupled multi-robot methods ~\cite{motes2023hypergraph} \rs{Slightly modifying your text. Add more exposition as necessary wrt theoretical guarantees you argue for the current paper. }{have been proposed. Hierarchical decomposition designed in this method can achieve planning efficiency, but introduces trade-offs in  probabilistic completeness.}.
% MR-TAMP with subtask dependencies~\cite{motes2020multi} \rs{I think the key point here is the monotone assumption}{formulates the high-level problem as a non-linear optimization and proposes an efficient rearrangement and action scheduling framework primarily focusing on monotone single-object interaction problems.}.

\section{Conclusion}
\label{sec:conclusion}

In this work, we formulate a general MR-TAMP problem as H-CSP when a task plan is given, which is inherently asynchronous. We propose a refinement planning algorithm driven by design principles and evaluate its efficiency and advantages over the synchronous approach in simulation. 

An immediate direction for future work is to develop a partial-order task planner capable of generating the input tuple of abstract actions and ordering constraints to complete the framework. This framework should facilitate bidirectional communication between the task planner and the proposed refinement planner to support full integration.

\bibliographystyle{IEEEtran}
\bibliography{IEEEabrv,references}

%\appendix
%\input{appendix}

\end{document}